\documentclass[conference]{IEEEtran}
\IEEEoverridecommandlockouts

\usepackage{cite}
\usepackage{amsmath,amsopn,amssymb}
\usepackage{graphicx,xspace,color,soul}
\usepackage{epsfig,subfigure}
\usepackage{longtable,multirow}
\usepackage{array,float}
\usepackage{url}
\usepackage{algorithm,algorithmicx,algpseudocode}
\usepackage{bm,epstopdf,booktabs}

\renewcommand{\vec}[1]{\mathbf{#1}}

\begin{document}

\title{Optimized Skeleton-based Action Recognition via \\ Sparsified Graph Regression}

\author{Xiang Gao, Wei Hu, Jiaxiang Tang, Jiaying Liu, Zongming Guo\\
Institute of Computer Science and Technology, Peking University, China\\
{\tt\small \{gyshgx868, forhuwei, hawkey1999, liujiaying, guozongming\}@pku.edu.cn}
}

\maketitle

\begin{abstract}
   With the prevalence of accessible depth sensors, dynamic human body skeletons have attracted much attention as a robust modality for action recognition. Previous methods model skeletons based on RNN or CNN, which has limited expressive power for irregular skeleton joints. While graph convolutional networks (GCN) have been proposed to address irregular graph-structured data, the fundamental graph construction remains challenging. In this paper, we represent skeletons naturally on graphs, and propose a graph regression based GCN (GR-GCN) for skeleton-based action recognition, aiming to capture the spatio-temporal variation in the data. As the graph representation is crucial to graph convolution, we first propose graph regression to statistically learn the underlying graph from multiple observations. In particular, we provide spatio-temporal modeling of skeletons and pose an optimization problem on the graph structure over consecutive frames, which enforces the sparsity of the underlying graph for efficient representation. The optimized graph not only connects each joint to its neighboring joints in the same frame strongly or weakly, but also links with relevant joints in the previous and subsequent frames. We then feed the optimized graph into the GCN along with the coordinates of the skeleton sequence for feature learning, where we deploy high-order and fast Chebyshev approximation of spectral graph convolution. Further, we provide analysis of the variation characterization by the Chebyshev approximation. Experimental results validate the effectiveness of the proposed graph regression and show that the proposed GR-GCN achieves the state-of-the-art performance on the widely used NTU RGB+D, UT-Kinect and SYSU 3D datasets.
\end{abstract}


\vspace{-0.1in}
\section{Introduction}

\begin{figure}[htbp]
    \centering
    \includegraphics[width=8.3cm]{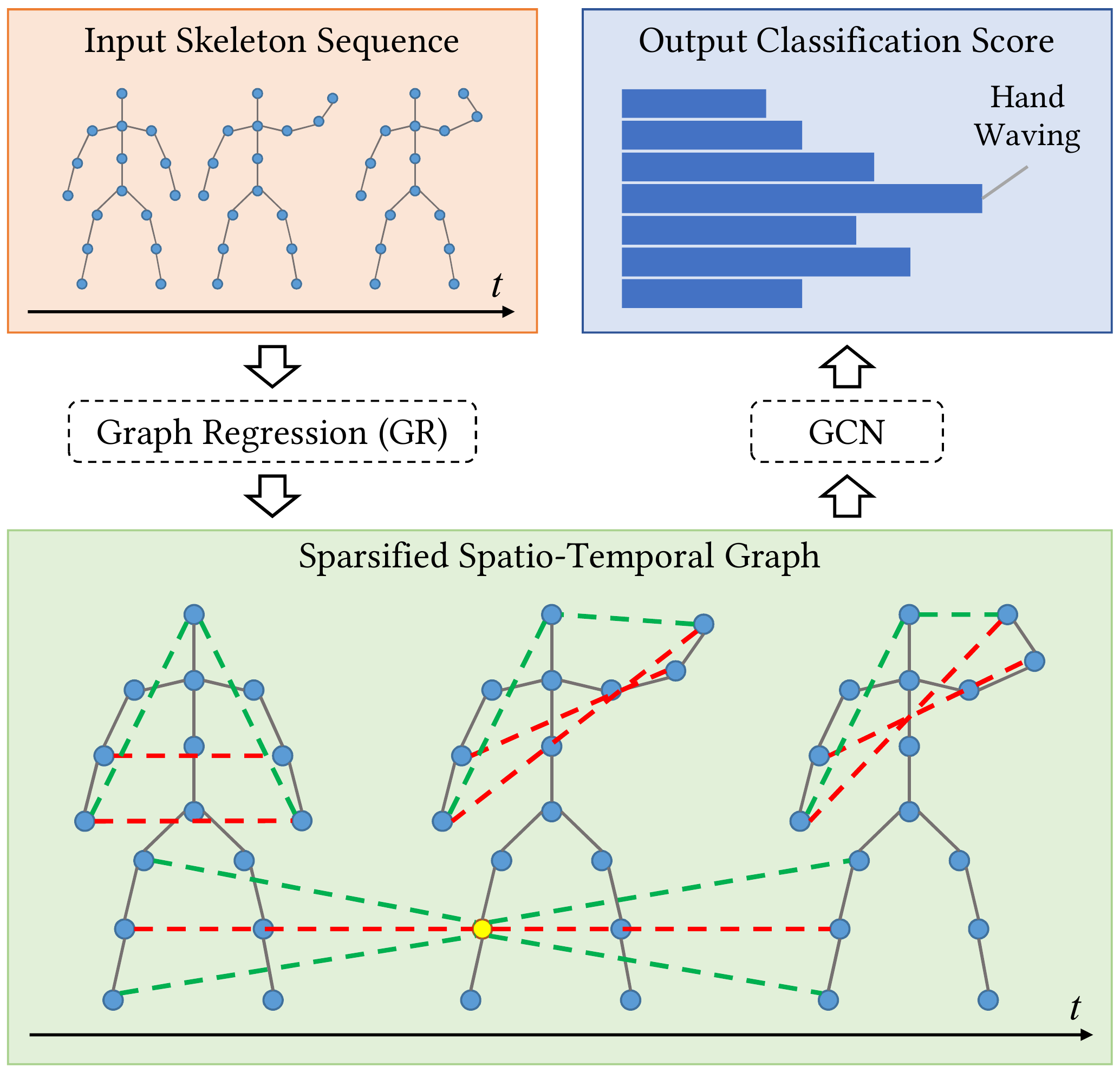}
    \caption{The pipeline of the proposed GR-GCN for skeleton-based action recognition. Given a sequence of human body joints, we first learn a common \textit{sparsified spatio-temporal graph} over each frame, its previous frame and the subsequent one via graph regression. This leads to a spatio-temporal graph with strong and physical edges (black solid lines), strong and non-physical edges (red dashed lines) and weak edges (green dashed ones) for \textit{variation modeling}. We then feed the sparsified spatio-temporal graph into a graph convolutional network (GCN) along with the 3D coordinates of joints for variation learning, which leads to the output classification scores.}
    \label{fig:teaser}
\end{figure}

Action recognition is an active research direction in computer vision, with widespread applications in video surveillance, human computer interaction, robot vision, autonomous driving and so on. Among the multiple modalities \cite{simonyan2014two,tran2015learning,wang2015action,wang2016temporal,zhao2017temporal} that are able to recognize human action, such as appearance, depth and body skeletons \cite{Du15cvpr,liu16eccv}, the skeleton-based sequences are springing up in recent years, due to the prevalence of affordable depth sensors (e.g., Kinect) and effective pose estimation algorithms \cite{shotton11}. Skeletons convey compact 3D position information of the major body joints, which are robust to variations of viewpoints, body scales and motion speeds \cite{Han17cviu}. Hence, skeleton-based action recognition has attracted more and more attention \cite{Xia12cvpr,Wang12cvpr,Gowayyed13IJCAI,Wang16eccv,vemulapalli14,Wang16cvpr,Weng17cvpr}. 

Different from modalities defined on regular grids such as images or videos, dynamic human skeletons are non-Euclidean geometric data, which consist of a series of human joint coordinates. This poses challenges in capturing both the intra-frame features and temporal dependencies. Recent methods learn these features via deep models like recurrent neural networks (RNN) \cite{Du15cvpr,liu16eccv,Shahroudy_2016_CVPR,Zhu16aaai,song2016aaai,zhang17,Li_2017_ICMEW,Tanfous_CVPR_2018,Liu_2018_TPAMI} and convolutional neural networks (CNN) \cite{Li_2017_ICMEW,li17,Ke17cvpr,kim2017interpretable,Liu17pr}. Nevertheless, the topology in skeletons is not fully exploited in the grid-shaped representation of RNN and CNN. 

A natural way to represent skeletons is graph, where each joint is treated as a vertex in the graph, and the relationship among the joints is interpreted by edges with weights. As unordered graphs cannot be fed into RNN or CNN directly, graph convolutional networks (GCN) have been proposed to deal with data defined on irregular graphs for a variety of applications \cite{bruna2013spectral,duvenaud2015convolutional,kipf2016semi,defferrard2016convolutional}. Yan et al. \cite{yan18} and Li et al. \cite{li18spatio} are the first to propose graph-based skeleton representation, which is then fed into the GCN to automatically learn the spatial and temporal patterns from data. Tang et al. \cite{Tang_2018_CVPR} propose a deep progressive reinforcement learning (DPRL) method to select the most informative frames of the input sequences and leverage GCN to learn the dependency among joints. Bin et al. \cite{bi2019spatio} propose a spatio-temporal graph routing (STGR) scheme for skeleton-based action recognition, which learns both the spatial connectivity and temporal connectivity. However, the graph constructions in these methods have certain limitations: graphs in \cite{yan18} are restricted by small partitions; graphs in \cite{li18spatio} only model joints bridged by a bone; there is no explicit temporal graph in \cite{Tang_2018_CVPR}; the computation complexity of graph learning in \cite{bi2019spatio} is high, and the spatial graph is built over clusters, each of which is assigned a weight and thus may not capture delicate pairwise spatial relationship among joints. 

Since the graph construction is crucial to graph convolution in GCNs, we propose a graph regression based GCN (GR-GCN) model to further improve the graph construction of skeleton data for stronger expressive power, providing an alternate view of the action sequence. The problem of learning the underlying graph structure from data (a.k.a., graph regression) is fundamental and helps discover the relation among graph signals. In the context of dynamic skeletons, we provide spatio-temporal modeling of skeletons and pose an optimization problem on the underlying graph Laplacian matrix\footnote{In spectral graph theory \cite{chung1997spectral}, a graph Laplacian matrix is an algebraic representation of the connectivities and node degrees of the corresponding graph, which will be introduced in Section \ref{sec:Laplacian}.} over consecutive frames. The optimization not only enforces the graph Laplacian to capture the structure of each spatio-temporal frame (i.e., every three consecutive frames), but also impose the sparsity constraint on the graph for compact representation. We then obtain the common structure of the graph Laplacian optimized from multiple observations of spatio-temporal frames by statistical analysis. The resulting graph not only connects each joint to its neighboring joints in the same frame strongly or weakly, but also links with relevant joints in the previous and subsequent frames. 

After learning the common optimal graph for spatio-temporal frames in a dynamic skeleton sequence, we feed the optimized graph into the GCN along with the coordinates of the skeleton sequence for feature learning. We deploy high-order and fast Chebyshev approximation of spectral graph convolution \cite{defferrard2016convolutional}, which leads to final classification scores. Further, we provide analysis of the variation characterization by the Chebyshev approximation. We analyze that the Chebyshev approximation essentially extracts the variation of the coordinates of joints, which is suitable to learn action features for final classification. As strong edges in the graph reflect strong relationship among physical/non-physical connections and weak edges represent potential relationship among non-physical connections, the proposed network strengthens learning actions which are accomplished by joints that are not bridged by bones (i.e., non-physical connections), such as ``drink water" with the interaction between one hand and the head.

In summary, our contributions include the following aspects:
\begin{itemize}
\setlength{\itemsep}{3pt}
\setlength{\parsep}{0pt}
\setlength{\parskip}{0pt}
    \item We propose efficient graph regression to learn the underlying common graph of spatio-temporal frames in a dynamic skeleton sequence, by posing an optimization problem on the graph Laplacian from the constraints of data structure and sparsity.  
    \item We integrate our graph regression with the GCN, and analyze the variation characterization by the Chebyshev approximation of spectral graph convolution, which leads to effective action feature learning.    
    \item We achieve the state-of-the-art performance on the widely used NTU RGB+D, UT-Kinect and SYSU 3D datasets, and validate the effectiveness of the proposed graph regression.
\end{itemize}

The rest of the paper is organized as follows. Section \ref{sec:related} reviews previous works on skeleton-based action recognition and GCNs. Next, we introduce some basic concepts in spectral graph theory in Section \ref{sec:Laplacian}. Then, we present the proposed spatio-temporal graph regression and sparsified graph construction in Section \ref{sec:modeling}, and elaborate on the proposed GR-GCN in Section \ref{sec:GR-GCN}. Finally, experimental results and conclusions are presented in Section \ref{sec:results} and Section \ref{sec:conclude}, respectively.

\begin{figure*}[t]
    \label{fig:framework}
    \centering
    \includegraphics[width=\textwidth]{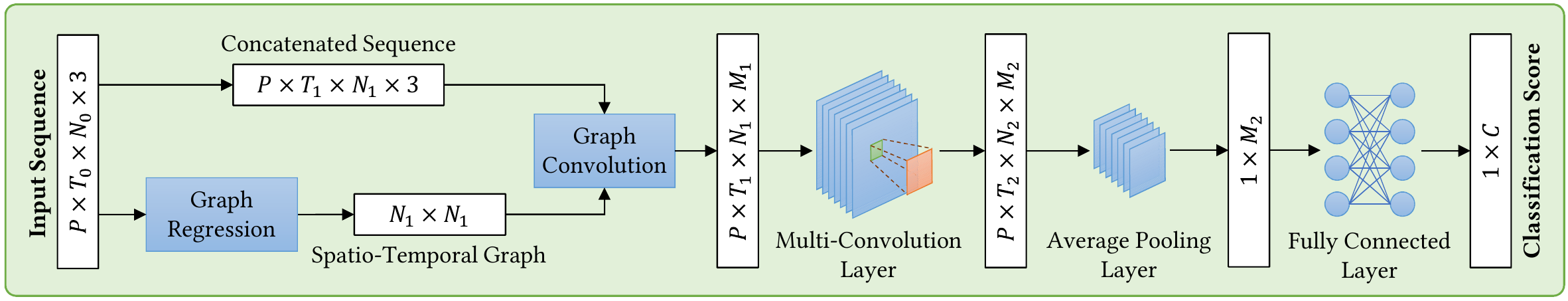}
    \caption{\textbf{The architecture of the proposed GR-GCN for skeleton-based action recognition.} Our proposed network takes a skeleton sequence as the input, which goes through sequence concatenation and sparsified spatio-temporal graph construction before feeding into the network. We then employ graph convolution and standard 2D convolution to the concatenated sequence, followed by feature aggregation via average pooling. Thereafter, a fully-connected layer is utilized to generate the output classification scores for $C$ classes.}
    \label{fig:pipeline}
\end{figure*}

\section{Related Work}
\label{sec:related}

\subsection{Skeleton-based Action Recognition} 
Previous skeleton-based action recognition methods can be divided into 2 classes \cite{yan18}: hand-crafted feature based methods and deep learning methods. 

\textbf{Hand-crafted feature based methods.} Hand-crafted features include covariance matrix for skeleton joint locations over time as a discriminative descriptor \cite{hussein13}, modeling human actions as curves in the Lie group \cite{vemulapalli14}, and Spatio-Temporal Naive-Bayes Nearest-Neighbor \cite{Weng17cvpr}, etc. However, these methods either lose information of interactions between specific sets of body parts or depend on complicated hand-crafted features.

\textbf{Deep learning methods.} Recent methods learn features via deep learning due to the notable performance, including RNN \cite{Du15cvpr,liu16eccv,Shahroudy_2016_CVPR,Zhu16aaai,song2016aaai,zhang17,Li_2017_ICMEW,Tanfous_CVPR_2018,Liu_2018_TPAMI} and CNN \cite{Li_2017_ICMEW,li17,Ke17cvpr,kim2017interpretable,Liu17pr}. However, these methods typically lose structural information when converting the raw skeleton data into the grid-shaped input of the neural networks. A natural way to address this issue is to represent skeleton data on graphs. Yan et al. \cite{yan18} and Li et al. \cite{li18spatio} are the first to employ GCNs to automatically learn both the spatial and temporal patterns from data. Specifically, Yan et al. \cite{yan18} construct graph convolution operations on partitions, which however may not capture the relationship among joints in different partitions due to the small receptive field. Li et al. \cite{li18spatio} design multi-scale convolutional filters, and simultaneously perform local convolutional filtering on temporal motions and spatial structures. For each frame, an undirected graph is constructed, where only joints bridged by a bone are connected, whereas there is no explicit temporal connectivity. Tang et al. \cite{Tang_2018_CVPR} propose a deep progressive reinforcement learning (DPRL) method to select the most informative frames of the input sequences and apply GCN to learn the spatial dependency between the joints. Edges in the constructed graph reflect both intrinsic dependencies (i.e., physical connection) and extrinsic dependencies (i.e., physical disconnection) by different weights. Nevertheless, there is no explicit graph construction in the temporal domain. Bin et al. \cite{bi2019spatio} propose a spatio-temporal graph routing (STGR) scheme for skeleton-based action recognition, which learns both spatial connectivity and temporal connectivity. Nevertheless, the computation complexity of the spatial and temporal graph learning is high. 

\subsection{Graph Convolutional Neural Networks}
GCN extends CNN by consuming data defined on irregular grids. The key challenge is to define convolution over graphs, which is difficult due to the unordered data. According to the definitions of graph convolution, most of these methods can be divided into two main categories: spectral-domain methods and nodal-domain methods.   

\textbf{Spectral-domain methods.} The convolution over graphs is elegantly defined in the spectral domain, which is the multiplication of the spectral-domain representation of signals. Specifically, the spectral representation is in the graph Fourier transform (GFT) \cite{hammond2011wavelets} domain, where each signal is projected onto the eigenvectors of the graph Laplacian matrix \cite{hammond2011wavelets, henaff2015deep}. The computation complexity, however, is high due to the eigen-decomposition of the graph Laplacian matrix in order to get the eigenvector matrix. Hence, it is improved by \cite{defferrard2016convolutional} through fast localized convolutions, where the Chebyshev expansion is deployed to approximate GFT. Besides, Susnjara et al. introduce the Lancoz method for approximation \cite{susnjara2015accelerated}. Spectral GCN has shown its efficiency in various tasks such as segmentation and classification \cite{kipf2016semi, Te18mm}.
    
\textbf{Nodal-domain methods.} Many techniques are introduced to implement graph convolution directly on each node and its neighbors, i.e., in the nodal domain. Gori et al. introduce recurrent neural networks that operate on graphs in \cite{gori2005new}. Duvenaud et al. propose a convolution-like propagation to accumulate local features \cite{duvenaud2015convolutional}. Bruna et al. deploy the multiscale clustering of graphs in convolution to implement multi-scale representation \cite{bruna2013spectral}. Furthermore, Niepert et al. define convolution on a sequence of nodes and perform normalization afterwards \cite{niepert2016learning}. Wang et al. propose edge convolution on graphs by incorporating local neighborhood information, which is applied to point cloud segmentation and classification \cite{wang2018dynamic}. Nodal-domain methods provide strong localized filters, but it also means it might be difficult to learn the global structure.

The above methods apply convolutional aggregators in the propagation step. Besides, there are other related works based on different aggregators, including attention aggregators \cite{velivckovic2018graph}, which incorporate the attention mechanism \cite{vaswani2017attention} into the propagation step, aiming to compute the hidden states of each node by attending over its neighbors; and gate aggregators \cite{li2016gated, tai2015improved, zayats2018conversation, peng2017cross, zhang2018sentence, liang2016semantic}, which use the gate mechanism like GRU \cite{cho2014association} or LSTM \cite{hochreiter1997long} in the propagation step to improve the long-term propagation of information across the graph structure.

\section{Preliminaries}
\label{sec:Laplacian}
We consider an undirected graph $ \mathcal{G}=\{\mathcal{V},\mathcal{E},\mathbf{A}\} $ composed of a vertex set $ \mathcal{V} $ of cardinality $|\mathcal{V}|=n$, an edge set $ \mathcal{E} $ connecting vertices, and a weighted \textit{adjacency matrix} $ \mathbf{A} $. $ \mathbf{A} $ is a real symmetric $ n \times n $ matrix, where $ a_{i,j} $ is the weight assigned to the edge $ (i,j) $ connecting vertices $ i $ and $ j $. We assume non-negative weights, \textit{i.e.}, $a_{i,j} \geq 0$.

\textit{The Laplacian matrix}, defined from the adjacency matrix, can be used to uncover many useful properties of a graph. Among different variants of Laplacian matrices, the \textit{combinatorial graph Laplacian} used in \cite{shen10pcs,hu14tip} is defined as 
\begin{equation}
   \mathbf{L}=\mathbf{D}-\mathbf{A}, 
   \label{eq:laplacian}
\end{equation}
where $ \mathbf{D} $ is the \textit{degree matrix}---a diagonal matrix where $ d_{i,i} = \sum_{j=1}^n a_{i,j} $. We will optimize $\mathbf{L}$ in the proposed graph regression method in Sec.~\ref{sec:modeling}. Further, the symmetric \textit{normalized} Laplacian is defined as $\mathcal{L} = \mathbf{D}^{-\frac{1}{2}}\mathbf{L}\mathbf{D}^{-\frac{1}{2}}$, which will be deployed in the GCN so as to avoid numerical instabilities. 



\textit{Graph signal} refers to data that resides on the vertices of a graph, such as social, transportation, sensor, and neuronal networks. In our context, we treat each joint in a skeleton sequence as a vertex in a graph, and define the corresponding graph signal as the coordinates of each joint.


\section{Dynamic Skeleton Modeling}
\label{sec:modeling}

The fundamental of skeleton-based action recognition is to capture the variation of joints both in the spatial and temporal domain, so as to learn motion features for classification. We propose spatio-temporal graph regression modeling for dynamic skeletons, and come up with the optimization of the underlying graph so as to characterize the variation.  

\subsection{Spatio-temporal Graph Regression Modeling of Skeletons}

Let $\vec{x}_t \in \mathbb{R}^{n \times 3}$ be the coordinate signal in one frame at time $t$, where $n$ is the number of joints in each skeleton. We define a \textit{spatio-temporal frame} as $\vec{x} = [\vec{x}_{t-1},\vec{x}_t,\vec{x}_{t+1}]^\top \in \mathbb{R}^{3n \times 3}$, i.e., three consecutive frames are concatenated. We then represent $\vec{x}$ on a spatio-temporal graph described by $\mathbf{L}$, which models the correlation among joints. 

We formulate the graph regression problem as the optimization of the graph Laplacian $\mathbf{L}$:
\begin{equation}
\begin{split}
    \min_{\mathbf{L}} \ \ &  \text{tr}(\vec{x}^\top \mathbf{L} \vec{x}) + \beta \| \mathbf{L} \|_F^2, \\
     \text{s.t.} \ \ & \text{tr}(\mathbf{L}) = 3n, \\
     & \mathbf{L}_{i,j}=\mathbf{L}_{j,i} \leq 0, i \neq j, \\
     & \mathbf{L} \cdot \vec{1}=\vec{0},
     \label{eq:skeleton_model}
\end{split}
\end{equation}
where $\beta$ is a weighting parameter, and $\vec{1}$ and $\vec{0}$ denote the constant one and zero vectors. In addition, $\text{tr}(\cdot)$ and $\| \cdot \|_F$ denote the trace and Frobenius norm of a matrix, respectively. The first term in the objective function aims to fit the graph structure to the data by minimizing the \textit{total variation} of the input signal (discussed soon), while the second term enforces the sparsity of the underlying graph for compact representation. The constraints ensure that the learned $\mathbf{L}$ satisfies the properties of the desired graph Laplacian: normalized, symmetry, non-negativity of edge weights, and the zero row sum. Next, we discuss the variation characterization by $\mathbf{L}$. 

The quadratic term $\vec{x}^\top \mathbf{L} \vec{x}$ in Eq.~(\ref{eq:skeleton_model}) describes the total variation. This is because $\vec{x}^\top \mathbf{L} \vec{x}$ can be written as \cite{yale04lect2}:
\begin{equation}
    \vec{x}^\top \mathbf{L} \vec{x} = \sum_{i \sim j} a_{i,j}(x_i-x_j)^2,
\end{equation}
where $i \sim j$ denotes two vertices $i$ and $j$ are one-hop neighbors in the graph. Hence, $\vec{x}^\top \mathbf{L} \vec{x}$ computes the total variation among connected vertices in $\vec{x}$. By minimizing this term in Eq.~(\ref{eq:skeleton_model}), we enforce the edge weight between a pair of vertices with different features to be small, while allowing for a large edge weight between a pair of similar vertices. Thus, the optimized graph is able to characterize the variation in the skeleton data.   

The optimization problem in Eq.~(\ref{eq:skeleton_model}) is convex and thus can be solved optimally, which leads to the learned graph for one given observation of $\vec{x}$. In order to acquire a spatio-temporal graph that captures the common structure of skeleton sequences, we propose to solve Eq.~(\ref{eq:skeleton_model}) over multiple observations of $\vec{x}$, and then statistically compute the common structure. For the purpose of succinct representation, we further restrict the connectivities of the common graph spatially and temporally, as discussed in the following.  

\subsection{Sparsified Graph Construction}
\label{subsec:graph}
The graph construction includes spatial connectivity and temporal connectivity. 

\textbf{Spatial connectivity.} Within each frame, we model the human body via a connected graph, based on two types of connectivities in particular: strong connections $\mathcal{E}_s$ and weak connections $\mathcal{E}_w$ for describing different correlations. Strong connections aim to capture strong correlations with large weights to emphasize the variation, including physical connectivity and some physical disconnection among joints, while weak connections are used to represent potential correlations among joints that are not physically connected. As shown in Fig.~\ref{fig:teaser}, whereas the ``head" joint and ``hand" joint are not bridged by a bone, a weak connectivity could be built between them because of the latent relationship during some actions (e.g., ``drink water"). In particular, different weights are assigned to strong and weak edges, i.e., edge weights within a frame are set as 
\begin{equation}
    a_{i,j} = \left \{
    \begin{array}{cc}
         w_1, & (i,j) \in \mathcal{E}_s  \\
         w_2, & (i,j) \in \mathcal{E}_w  \\
         0, & \text{otherwise},
    \end{array}
    \right.
    \label{eq:inter_weight}
\end{equation}
where $w_1 > w_2$. 

\begin{figure}[t]
    \centering
    \includegraphics[width=0.45\textwidth]{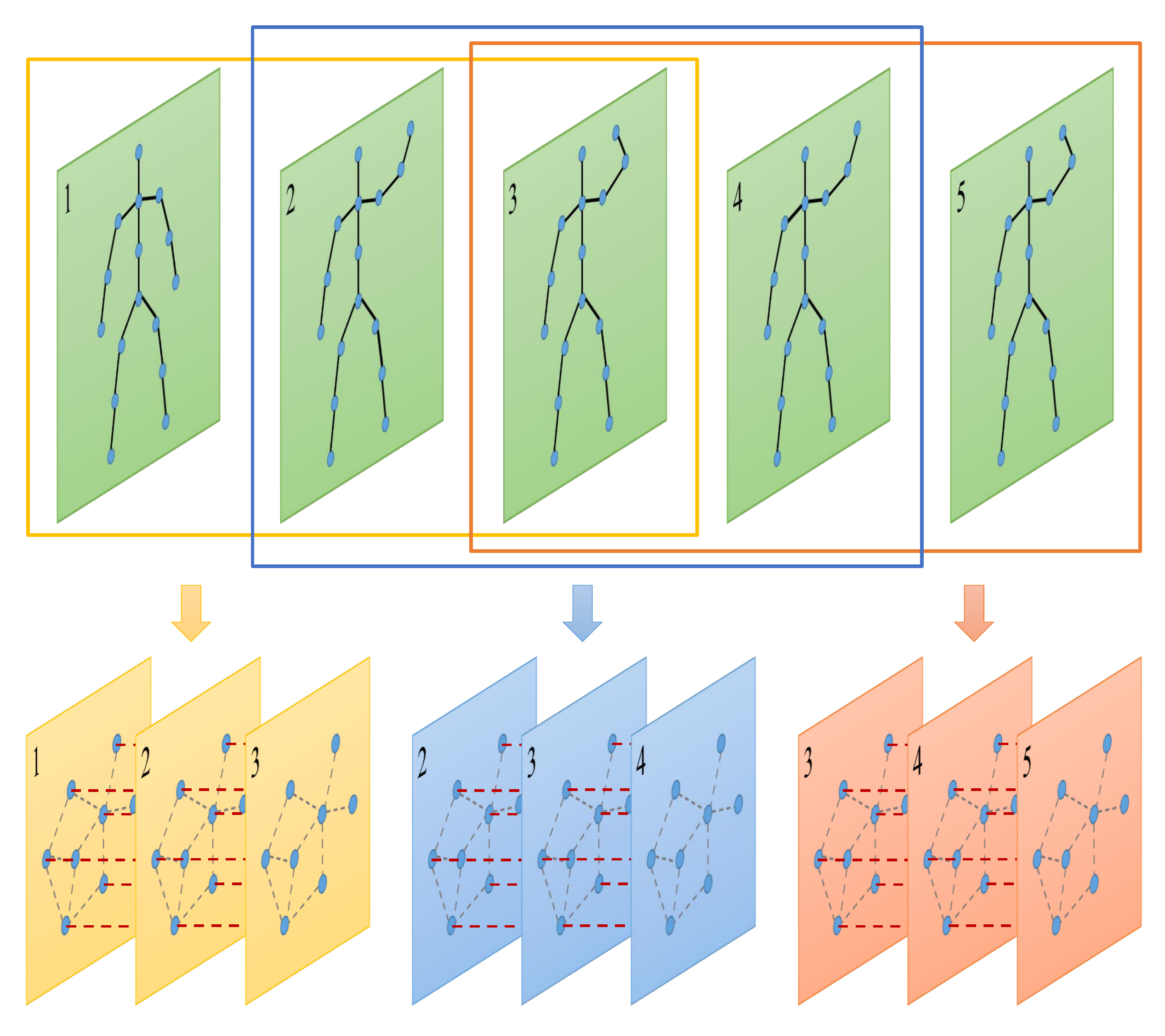}
    \caption{\textbf{Illustration of the learned graph construction.} We learn a common structure of the spatio-temporal graph for  spatio-temporal frames (i.e., every three adjacent frames). The yellow, blue, and red groups include three adjacent frames respectively, containing intra-frame connectivities (gray dotted lines) and inter-frame connectivities (red dotted lines). Note that the connectivities are simplified for clear visualization.}
    \label{fig:graph}
\end{figure}

\textbf{Temporal connectivity.} Unlike previous works where each joint is disconnected in the temporal domain or only connected to its corresponding joints in the adjacent frames in general, we allow connecting each joint in frame $\mathbf{x}_t$ to the neighborhood of its correspondence in the previous frame $\mathbf{x}_{t-1}$ and subsequent frame $\mathbf{x}_{t+1}$, which are referred to as \textit{potential edges}, as shown in Fig.~\ref{fig:graph}. This is to capture the latent variation between one joint in frame $\mathbf{x}_t$ and its neighboring joints in the adjacent frames. The receptive field in the temporal domain is thus enlarged by exploiting more neighboring joints, which contributes to learning the temporal variation. Taking the action ``typing on a keyboard" as an example, the left thumb may have little motion in a short period. However, the left index finger moves relative to the left thumb both spatially and over time, which can be captured by the proposed potential edges. Hence, the final spatio-temporal adjacency matrix of consecutive frames $\{\mathbf{x}_{t-1},\mathbf{x}_t,\mathbf{x}_{t+1}\}$ is defined as 
\begin{equation}
    \mathbf{A}_g = 
    \begin{bmatrix}
    \mathbf{A}_{t-1,t-1} & \mathbf{A}_{t-1,t} & \mathbf{O} \\
    \mathbf{A}_{t,t-1} & \mathbf{A}_{t,t} & \mathbf{A}_{t,t+1} \\
    \mathbf{O} & \mathbf{A}_{t+1,t} & \mathbf{A}_{t+1,t+1} 
    \end{bmatrix},
\end{equation}
where $\mathbf{O} \in \mathbb{R}^{n \times n}$ is a zero matrix, $\mathbf{A}_{i,i} \in \mathbb{R}^{n \times n}$ is the weighted adjacency matrix of frame $i$ for representing the intra-frame connectivity, while $\mathbf{A}_{i,j} \in \mathbb{R}^{n \times n} (i \neq j)$ is the adjacency matrix between frame $i$ and $j$ for description of the inter-frame connectivity. 

Similar with the edge weights for the spatial connectivity, we define two types of temporal connectivities: the connectivity for corresponding joints, denoted as $\mathcal{E}_c$, and the connectivity between each joint and the neighborhood of its correspondence in the adjacent frames, denoted as $\mathcal{E}_n$. We assign $w_1$ to edges in $\mathcal{E}_c$, and assign $w_2$ to edges in $\mathcal{E}_n$, i.e., 
\begin{equation}
    a_{i,j} = \left \{
    \begin{array}{cc}
         w_1, & (i,j) \in \mathcal{E}_c  \\
         w_2, & (i,j) \in \mathcal{E}_n  \\
         0, & \text{otherwise},
    \end{array}
    \right.
    \label{eq:intra_weight}
\end{equation}
where $i$ and $j$ denote vertices in two different frames.

\subsection{Final Graph Modeling}
Based on the above restriction of spatial and temporal connectivities, we extract the common structure of the optimized graph Laplacian learned from multiple observations of spatio-temporal frames. Specifically, we first randomly take $m$ spatio-temporal frames from different classes of skeleton sequences, each of which serves as $\mathbf{x}$ in Eq.~(\ref{eq:skeleton_model}). Then we obtain the optimal spatio-temporal graph Laplacian for each spatio-temporal frame Eq.~(\ref{eq:skeleton_model}), leading to $m$ optimized graph Laplacian $\{\mathbf{L}_{\text{opt}}^l\}_{l=1}^m$. Next, we derive a common graph Laplacian $\mathbf{L}$ from the statistics of $\{\mathbf{L}_{\text{opt}}^l\}_{l=1}^m$.

\section{The Proposed GR-GCN}
\label{sec:GR-GCN}
Having elaborated on the proposed graph regression that provides the underlying common structure of spatio-temporal frames, we now overview the architecture of the proposed GR-GCN. Then we discuss the corresponding graph convolution and feature learning in detail.

\subsection{GR-GCN architecture}

As illustrated in Fig.~\ref{fig:pipeline}, the input is a skeleton-based action sequence organized as a $P \times T_0 \times N_0 \times 3$ tensor, where $P$ is the number of actors in each sequence, $T_0$ is the number of frames, $N_0$ is the number of joints in each frame, and $3$ means the dimension of $x$, $y$, $z$ coordinates. In order to exploit the spatio-temporal dependencies, we firstly concatenate the input sequence in the unit of $3$ consecutive frames, e.g., the $\{1,2,3\}^\text{th}$ frames are concatenated into the first spatio-temporal frame, and the $\{2,3,4\}^\text{th}$ frames into the second one, etc. Thus, the sequence length is changed to $T_1$, and the number of joints in each frame is $N_1$ after frame concatenation, where $T_1=T_0-2$ and $N_1=N_0 \times 3$. We then perform the proposed graph regression, which leads to the learned graph Laplacian of a common spatio-temporal graph. Secondly, we feed a feature matrix containing the coordinates of skeleton joints in the concatenated sequence and the graph Laplacian into the designed graph convolution layer and standard 2D convolution layers for feature extraction. Average pooling is then employed for feature aggregation. Finally, the global feature matrix will go through a fully connected layer followed by a Softmax activation function to output the classification score for $C$ classes. Also, batch normalization is used for all layers before the ReLU activation function.

\subsection{Spatio-Temporal Graph Convolution}
\label{subsec:convolution}
Following the definition of graph convolution in \cite{defferrard2016convolutional}, we adopt the approximation of spectral convolution by Chebyshev polynomials for efficient implementation: 
\begin{equation}
    g_{\theta} \ast \mathbf{x} \approx \sum_{k=0}^{K-1} \theta_k T_k(\mathcal{L})\mathbf{x},
 \label{eq:chebyshev}
\end{equation}
where $\mathcal{L}=\mathbf{D}^{-\frac{1}{2}}\mathbf{L}\mathbf{D}^{-\frac{1}{2}}$ is the symmetric normalized graph Laplacian as defined in Sec.~\ref{sec:Laplacian}, which is employed because the domain of Chebyshev polynomials lies in $[-1,1]$. $\theta_k$ denotes the $ k $-th Chebyshev coefficient and $g_{\theta}$ denotes a convolution kernel. $T_k(\mathcal{L})$ is the Chebyshev polynomial of order $ k $. It is recurrently calculated by $ T_k(\mathcal{L}) = 2\mathcal{L}T_{k-1}(\mathcal{L}) - T_{k-2}(\mathcal{L})$, where $T_0(\mathcal{L}) = 1,T_1(\mathcal{L}) = \mathcal{L}$. When $k > 1$, $\mathcal{L}^k$ essentially describes $k$-hop connectivities, thus incorporating more neighbors and leading to convolution over a larger receptive field.     

We provide analysis of the variation characterization by the above Chebyshev approximation. As discussed in \cite{Shuman2013The}, the graph Laplacian matrix $\mathbf{L}$ is essentially a high-pass operator which captures the variation in the underlying signal. For any signal $\mathbf{x}$, it satisfies
\begin{equation}
    (\mathbf{L}\mathbf{x})(i) = \sum_{j \in \mathcal{N}_i} a_{i,j} (x_i - x_j),
    \label{eq:model}
\end{equation}
where $(\mathbf{L}\mathbf{x})(i)$ denotes the $i$-th component of $\mathbf{L}\mathbf{x}$. $\mathcal{N}_i$ is the set of vertices connected to $i$. This presents that when operating $\mathbf{L}$ on $\mathbf{x}$, for each vertex, it computes the signal difference among the vertex and its one-hop neighbors. In other words, $\mathbf{L}\mathbf{x}$ captures the variation in $\mathbf{x}$. Similarly, $\mathbf{L}^k\mathbf{x}$ captures the variation between each vertex and its $k$-hop neighbors. Thus, the approximated graph convolution seamlessly enables learning the variation in a skeleton sequence. This also sheds light on why graph convolution works for action recognition.

\subsection{Feature Learning}
Having designed the spatio-temporal graph convolution, we define the transfer function as follows:
\begin{equation}
    \mathbf{y} = \text{ReLU}(\sum_{k=0}^{K-1} T_k(\mathcal{L})\mathbf{x}\mathbf{W}_k + \mathbf{b}),
    \label{eq:featureLearning}
\end{equation}
where $\mathbf{W}_k \in \mathbb{R}^{F_1 \times F_2}$ is a matrix of weight parameters $\theta_k$ as in Eq.~\ref{eq:chebyshev}, which will be learnt from the network, and $F_1$, $F_2$ are the dimensions of generated features in two connected layers respectively. $\mathbf{b} \in \mathbb{R}^{n \times F_2}$ is the bias, while ReLU is an activation function. 


After the graph convolution layer, we employ standard 2D convolution to the output $\mathbf{y}$, followed by feature aggregation via average pooling. Thereafter, a fully-connected layer and a Softmax activation function are adopted to generate the output classification scores. We adopt the categorical cross-entropy loss to train the network. The implementation details of our model will be discussed in Sec.~\ref{subsec:implement}.

\section{Experiments}
\label{sec:results}

We evaluate our proposed GR-GCN on four widely used datasets and compare with state-of-the-art skeleton-based action recognition methods. Experimental details and results are discussed below.

\subsection{Datasets and Evaluation Metrics}

\textbf{NTU RGB+D Dataset} \cite{Shahroudy_2016_CVPR}: This dataset was captured from 40 human subjects by 3 Microsoft Kinect v2 cameras. It consists of 56880 action sequences with 60 classes. Actions 1-49 were performed by one actor, and actions 50-60 were performed by the other two actors. Each body skeleton was recorded with 25 joints. The benchmark evaluations include Cross-Subject (CS) and Cross-View (CV). In the CS evaluation, 40320 samples from 20 subjects were used for training, and the other samples for testing. In the CV evaluation, samples captured from camera 2 and 3 were used for training, while samples from camera 1 were employed for testing.

\textbf{Florence 3D Dataset} \cite{Lorenzo_CVPRW_2013}: This dataset contains 215 action sequences of 10 actors with 9 classes. Each body skeleton was collected from Kinect, and recorded with 15 joints. We follow the standard experimental settings to perform leave-one-actor-out validation protocol: we use all the sequences from 9 out of 10 actors for training and the remaining one for testing, and repeat this procedure for all the actors. The resulting 10 classification accuracy values are averaged to get the final accuracy.

\textbf{UT-Kinect Dataset} \cite{Xia12cvpr}: This dataset was captured using a single stationary Kinect. It consists of 200 sequences with 10 classes, and each skeleton includes 20 joints. The dataset was recorded by three channels: RGB, depth, and skeleton joint locations, whereas we only use the 3D skeleton joint coordinates. We also adopt the leave-one-actor-out validation protocol to evaluate our model on this dataset.

\textbf{SYSU 3D Dataset} \cite{Hu_2015_CVPR}: On this dataset, 40 actors were asked to perform 12 different activities. Therefore, there are totally 480 action videos on this dataset. For each video, the corresponding RGB, depth, and skeleton information were captured by a Kinect. We use the skeleton sequences performed by 20 actors for training, and the remaining 20 actors for testing. We employ the 30-fold cross-subject validation and report the mean accuracy on the dataset.

\subsection{Implementation Details}
\label{subsec:implement}

Our proposed model was implemented with the PyTorch\footnote{https://pytorch.org} framework. The number of actors $P$ is set to be 2, 1, 1, 1 for NTU RGB+D, Florence 3D, UT-Kinect, and SYSU 3D dataset respectively. We learn the edge weight ratio $r=5$ for the four datasets, i.e., $w_1 = 5$, $w_2 = 1$. 

\textbf{Basic Model}: Prior to the graph convolution layer, we set a Batch Normalization layer for the batched input data in order to be less careful about data initialization and speed up the training process \cite{Sergey_2015_ICML}. In the graph convolution layer, we set the Chebyshev order $K$ to be 4, and the dimension of the weight matrix $\mathbf{W}_k$ in Eq.~\ref{eq:featureLearning} to be $3n \times 3n$ (i.e., the same as the spatio-temporal Laplacian matrix $\mathcal{L}$). The Multi-Convolution Layer consists of 2 standard CNN layers. Each convolution layer follows a Batch Normalization layer. We choose ReLU as the activation function after each convolution layer, and assign the dropout rate 0.5.

\textbf{Deep Stacking}: The above convolutional model can be easily extended into a deep architecture. Taking the above model as one basic layer, we stack it into a multi-layer network architecture, in which the output at the previous layer is used as the input of the next layer. Here, we stack it into a 10-layer architecture. In this architecture, we appropriately
adjust the kernel size so as to acquire the final output feature of dimension $M_2=256$ for each point. With the increase of layers, the receptive field of convolutional kernels become larger, thus enabling abstracting more global information.

Next, we employ three average pooling layers to pool the $P$, $N$, and $T$ dimension respectively, followed by a fully connected layer and a Softmax activation function to output the final classification score. The number of neurons depends on the output channel of the last convolution layer of the network. We apply Adam \cite{kingma2014adam} optimizer to train the whole model with the initial learning rate 0.1, and decrease it on the 10$^\text{th}$ epoch. Note that we did not perform any normalization on the skeleton coordinates during data preprocessing.

\subsection{Data Preprocessing}
\label{subsec:preprocess}

\textbf{NTU RGB+D Dataset:} Due to some missing skeletons in this dataset, we only use the cleaned data\footnote{https://github.com/InwoongLee/TS-LSTM} for action recognition \cite{lee17}. In order to enhance the robustness of model training, we split the sequences into several segments of equal size in a way similar to \cite{li18spatio}. Specifically, we split the whole sequence into 32 segments, and pick the \{1, 2, 3, 4\}$^\text{th}$ frame respectively from each segment to generate a large amount of training data.
    
    
\textbf{Florence 3D Dataset:} Since the sequences in this dataset contain few frames, we design two ways to generate the training set: sampling and interpolation. For longer sequences (i.e., the length of the sequence is greater than 32), we randomly choose 32 frames; for the other sequences, we calculate the mean of two adjacent frames and insert it into the sequence as a new frame, eventually forming a sequence of 32 frames. For all the sequences, we repeat this operation 3 times to generate the training set.
    

\textbf{UT-Kinect Dataset:} We also adopt sampling and interpolation methods to generate the training set. Here, we set the length of each training sequence to be 64, and repeat the process twice. 
    
\textbf{SYSU 3D Dataset:} Similar to the NTU RGB+D dataset, we split each sequence into 32 segments, and pick the \{1, 2, 3, 4, 5\}$^\text{th}$ frame from each segment to generate the training set. However, this dataset does not provide vertex labels, hence we only adopt the adjacency matrix of physical connections provided by the author as the graph within each frame.
    

\subsection{Results on NTU RGB+D Dataset}
\label{subsec:ntu}
As reported in Tab.~\ref{tb:ntu}, our model achieves accuracy of 87.5\% in CS and 94.3\% in CV respectively. Also, as will be discussed in the ablation study, the proposed intra-connections improve the performance by 0.7\% in CS and 1.4\% in CV over the baseline method (\textit{GR-GCN+Bone}), while the proposed temporal connectivities lead to 3.2\% gain in CS and 3.1\% gain in CV, thus validating the effectiveness of our method.   

\textbf{Comparison with the State-of-the-arts:} We present the comparison with the state-of-the-art methods in Tab.~\ref{tb:ntu}. We see that our method outperforms all the other state-of-the-art methods. Specifically, compared with the latest state-of-the-art method STGR-GCN \cite{bi2019spatio}, our model leads to 0.6\% gain in CS and 2.0\% gain in CV respectively, which demonstrates the superiority of our method.


\textbf{Ablation Study:} In order to validate the advantages of the proposed spatio-temporal graph construction in our method, we evaluate various graph construction methods progressively and design the following incomplete models. Model 1 is \textit{GR-GCN (Bone only)}, in which only joints connected with a bone are linked with graph edges. This kind of graph construction is commonly used in existing graph-based skeleton recognition \cite{yan18,li18spatio,Tang_2018_CVPR}, and thus is the baseline. Model 2 is \textit{GR-GCN (Bone + Intra-connection)} (non-physical), where connectivities are further added to joints that are not physically connected within each frame, including strong and weak edges for capturing latent dependencies. This kind of connectivities are previously exploited in \cite{Tang_2018_CVPR}. Model 3 is our complete model with extra temporal connections included. We observe that Model 1 already achieves competitive performance with the state-of-the-art methods, which shows the effectiveness of the proposed GR-GCN. With additional intra-connectivities, Model 2 improves the accuracy by 0.7\% in CS and 1.4\% in CV over Model 1, validating the benefits of non-physical connections. Further, when the temporal connections are exploited, the complete model achieves 2.5\% gain in CS and 1.7\% gain in CV over Model 2. We thus conclude that both the proposed non-physical intra-connectivities and the explicit temporal connections make contributions to skeleton-based action recognition, in which the temporal connectivities are more crucial.

\begin{table}
  \centering
  \caption{Comparisons on the NTU RGB+D dataset (\%).}
  \label{tb:ntu}
  \begin{tabular}{rccc}
    \hline
     \textbf{Methods} & \textbf{CS} & \textbf{CV} & \textbf{Year} \\
    \hline
    Dynamic Skeletons \cite{Hu_2015_CVPR} & 60.2 & 65.2 & 2015 \\
    Part-aware LSTM \cite{Shahroudy_2016_CVPR} & 62.9 & 70.3 & 2016 \\
    Geometric Features \cite{zhang17}  & 70.3 & 82.4 & 2017 \\
    LSTM-CNN \cite{Li_2017_ICMEW} & 82.9 & 91.0 & 2017 \\
    Two-Stream CNN \cite{li17}  & 83.2 & 89.3 & 2017 \\
    ST-LSTM (Tree)+Trust Gate \cite{Liu_2018_TPAMI} & 69.2 & 77.7 & 2018 \\
    Deep STGC$_K$ \cite{li18spatio}  & 74.9 & 86.3 & 2018 \\
    ST-GCN \cite{yan18}  & 81.5 & 88.3 & 2018 \\
    DPRL \cite{Tang_2018_CVPR}  & 83.5 & 89.8 & 2018 \\
    SR-TSL \cite{si18}  & 84.8 & 92.4 & 2018 \\
    STGR-GCN \cite{bi2019spatio} & 86.9 & 92.3 & 2019 \\
    \hline
    GR-GCN (Bone only) & 84.3 & 91.2 &  \\
    GR-GCN (Bone + Intra-connection) & 85.0 & 92.6 & \\
    Complete GR-GCN model & \textbf{87.5} & \textbf{94.3} & \\
    \hline
  \end{tabular}
\end{table}


\subsection{Results on SYSU 3D Dataset}

We compare our method with the state-of-the-art skeleton-based action recognition  methods on SYSU 3D Dataset, which are presented in Tab.~\ref{tb:sysu}. Our proposed method outperforms all the other state-of-the-art methods on this dataset, achieving accuracy improvement of 1.0\% over the previous best method DPRL \cite{Tang_2018_CVPR}. 

Note that, as vertex labels are not provided by this dataset, we can only build strong physical connections from the given adjacency matrix within each frame while abandoning weak edges. Hence, we provide ablation study with Model 1 in Tab.~\ref{tb:sysu}. We see that our complete model achieves 2.7\% improvement over the baseline method. This validates the benefits of incorporating explicit temporal connectivities across consecutive frames again. 

\begin{figure}[htbp]
    \centering
    \includegraphics[width=0.45\textwidth]{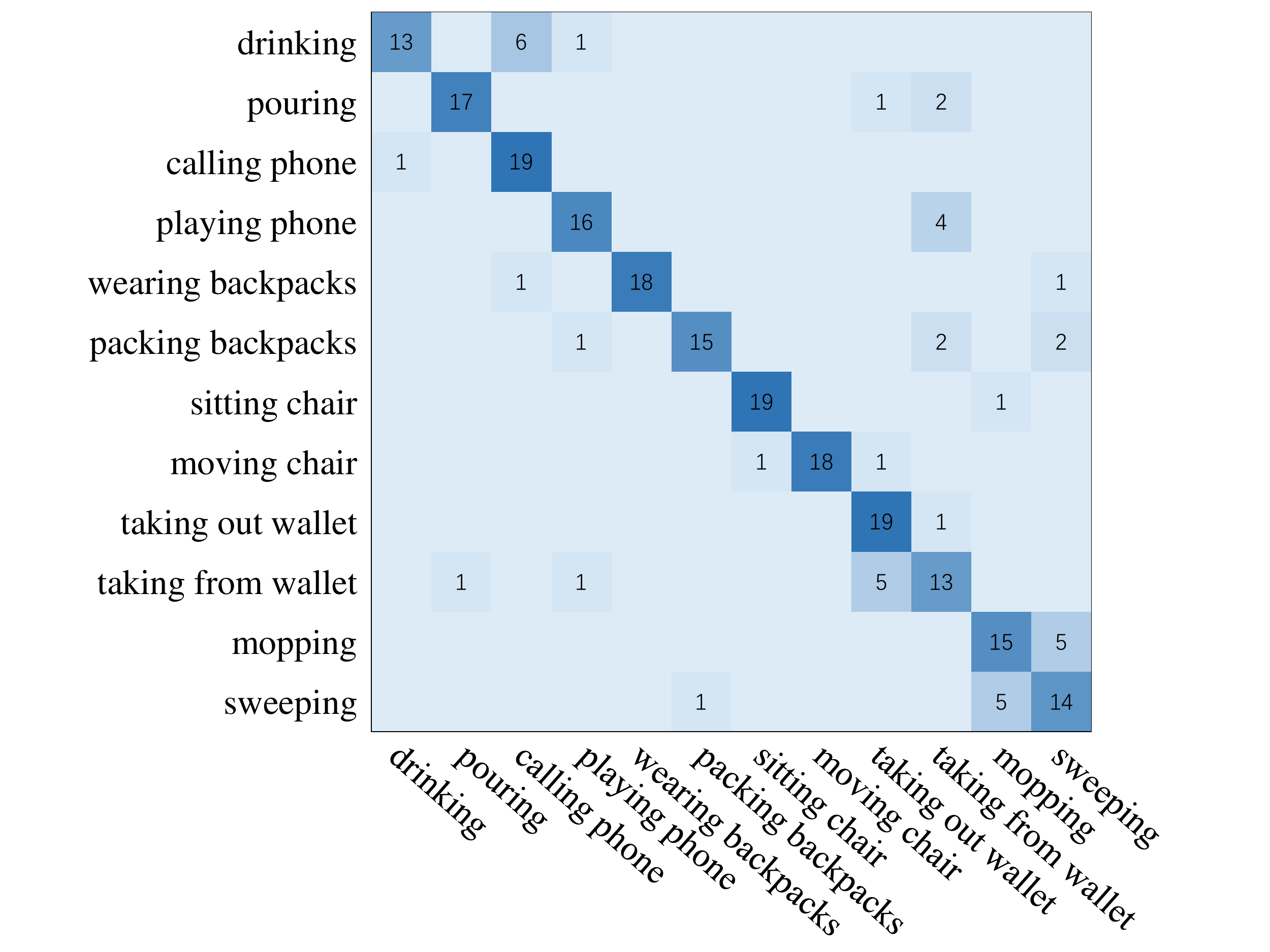}
    \caption{Confusion matrix of GR-GCN on SYSU 3D dataset.}
    \label{fig:cmat}
\end{figure}

Also, the confusion matrix of our result is demonstrated in Fig.~\ref{fig:cmat}. We see that the matrix is diagonally dominant on all the 12 classes, which validates that our method achieves excellent classification results on this dataset. Besides, we note that our model sometimes confuses the activity of ``mopping" with ``sweeping", which is mainly due to the highly similar motions in the two actions.

\begin{table}
  \centering
  \caption{Comparisons on the SYSU 3D dataset (\%).}
  \label{tb:sysu}
  \begin{tabular}{rcc}
    \hline
     \textbf{Methods} & \textbf{Accuracy} & \textbf{Year} \\
    \hline
    Dynamic Skeletons \cite{Hu_2015_CVPR} & 75.5 & 2015 \\
    LAFF (SKL) \cite{hu2016real} & 54.2 & 2016 \\
    ST-LSTM (Tree) \cite{Liu_2018_TPAMI} & 73.4 & 2018 \\
    ST-LSTM (Tree) + Trust Gate \cite{Liu_2018_TPAMI} & 76.5 & 2018 \\
    DPRL \cite{Tang_2018_CVPR}  & 76.9 & 2018 \\
    \hline
    GR-GCN (Bone only) & 75.2 &  \\
    Complete GR-GCN model & \textbf{77.9} & \\
    \hline
  \end{tabular}
\end{table}

\subsection{Results on UT-Kinect Dataset}

As listed in Tab.~\ref{tb:ut}, our method achieves comparable accuracy of 98.5\% to \cite{Tang_2018_CVPR}, and outperforms all the other methods. Note that the performance difference among all the methods is rather small in general. The reason is that this dataset includes several very similar actions, which are difficult to distinguish without RGB or depth data.

Also, we perform the same ablation study as in Sec.~\ref{subsec:ntu}, as reported in Tab.~\ref{tb:ut}. We observe that Model 2 improves the accuracy by 0.5\% over Model 1 with additional intra-connectivities. Further, when the temporal connectivities are built, the complete model achieves 1.1\% improvement over Model 2, which demonstrates the advantages of the proposed spatio-temporal graph construction.   

\begin{table}
  \centering
  \caption{Comparisons on the UT-Kinect dataset (\%).}
  \label{tb:ut}
  \begin{tabular}{rcc}
    \hline
     \textbf{Methods} & \textbf{Accuracy} & \textbf{Year} \\
    \hline
    Lie Group \cite{vemulapalli14} & 97.1 & 2014 \\
    LARP+mfPCA \cite{Rushil_2015_CVPR} & 94.9 & 2015 \\
    SPGK \cite{Wang16eccv} & 97.4 & 2016 \\
    ST-NBNN \cite{Weng17cvpr} & 98.0 & 2017 \\
    Bi-LSTM \cite{Tanfous_CVPR_2018} & 96.9 & 2018 \\
    ST-LSTM(Tree) + Trust Gate \cite{Liu_2018_TPAMI} & 97.0 & 2018 \\
    DPRL \cite{Tang_2018_CVPR}  & 98.5 & 2018 \\
    \hline
    GR-GCN (Bone only) & 96.9 &  \\
    GR-GCN (Bone + Intra-connection) & 97.4 & \\
    Complete GR-GCN model & \textbf{98.5} & \\
    \hline
  \end{tabular}
\end{table}

\subsection{Results on Florence 3D Dataset}

We present the performance comparison with the state-of-the-art methods on the Florence 3D dataset in Tab.~\ref{tb:florence}. Our method achieves classification accuracy of 98.5\%, outperforming all the other state-of-the-art methods significantly except Deep STGC$_K$ \cite{li18spatio}. The reason is that Deep STGC$_K$ benefits from the design philosophy of autoregressive moving average model, which is tailored for time sequences. Due to the few joints in each frame and few frames in the sequence, our model is difficult to capture subtle variation from few joints. Thus we misclassify ``drink from a bottle" and ``answer phone", ``read watch" and ``clap", which is difficult to distinguish even with human vision.

Moreover, Tab.~\ref{tb:florence} reports the results of ablation study. We achieve 0.1\% improvement from non-physical intra-connections compared with GR-GCN (Bone only), and another 2.8\% improvement from temporal connections compared with GR-GCN (Bone + Intra-connection). This validates the effectiveness of the proposed graph construction, in which the temporal connectivities are vital. 

\begin{table}
  \centering
  \caption{Comparisons on the Florence 3D dataset (\%).}
  \label{tb:florence}
  \begin{tabular}{rcc}
    \hline
     \textbf{Methods} & \textbf{Accuracy} & \textbf{Year} \\
    \hline
    Lie Group \cite{vemulapalli14} & 90.9 & 2014 \\
    LARP+mfPCA \cite{Rushil_2015_CVPR} & 89.7 & 2015 \\
    Rolling Rotations \cite{Raviteja_2016_CVPR} & 91.4 & 2016 \\
    SPGK \cite{Wang16eccv} & 91.6 & 2016 \\
    Transion Forests \cite{Guillermo_2017_CVPR} & 94.2 & 2017 \\
    MIMTL \cite{yang2017} & 95.3 & 2017 \\
    Bi-LSTM \cite{Tanfous_CVPR_2018} & 93.0 & 2018 \\
    Deep STGC$_K$ \cite{li18spatio}  & \textbf{99.1} & 2018 \\
    \hline
    GR-GCN (Bone only) & 95.5 &  \\
    GR-GCN (Bone + Intra-connection) & 95.6 & \\
    Complete GR-GCN model & 98.4 & \\
    \hline
  \end{tabular}
\end{table}

\begin{figure}[htbp]
    \centering
    \includegraphics[width=0.45\textwidth]{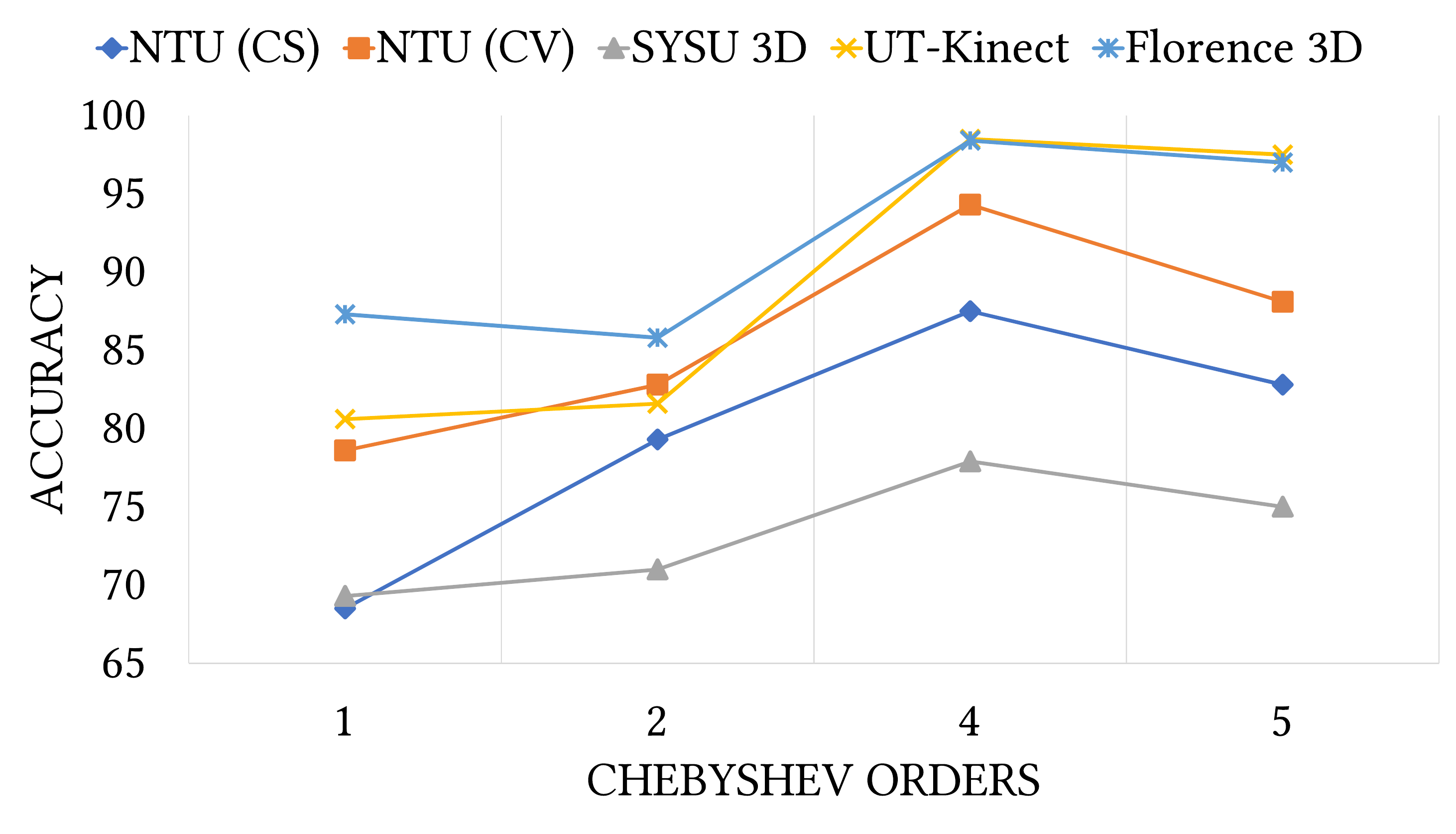}
    \caption{Classification accuracy on different Chebyshev orders.}
    \label{fig:cheby_order}
\end{figure}

\subsection{Analysis on Chebyshev Orders}

We explore the effects of different Chebyshev polynomial orders on our complete GR-GCN model, as demonstrated in Fig.~\ref{fig:cheby_order}.
When $K=1$, graph convolution defaults to a fully connected layer according to Eq.~(\ref{eq:chebyshev}), thus becoming the baseline with only traditional CNNs. The performance is inferior to those with larger $K$ (corresponding to graph convolution with $(K-1)$-hop neighborhood) in general, thus validating the effectiveness of graph convolution. Further, our model achieves the best performance when $K=4$  for all the datasets, thus validating the choice of $K$ in the experimental setting. In contrast, the performance with $K=5$ drops, because a wide range of neighbors will be incorporated, which is unable to capture the local variation well and may lead to overfitting.

\section{Conclusion}
\label{sec:conclude}

We propose a graph regression based GCN (GR-GCN) for skeleton-based action recognition, aiming to fully exploit both spatial and temporal dependencies among human joints. As the graph representation is crucial to graph convolution, we propose graph regression to optimize the underlying graph over multiple observations of spatio-temporal frames, and then statistically learn the common sparsified graph representation. The learned graph not only captures intrinsic physical connections, but also models non-physical spatial connectivities as well as temporal connectivities over consecutive frames so as to represent the latent correlations for better action recognition. We then feed the learned spatio-temporal graph into the GCN with spectral graph convolution approximated by high-order Chebyshev polynomials for feature extraction. Extensive experiments demonstrate the superiority of our method.


\end{document}